\documentclass[10pt, letter,twocolumn]{IEEEtran}
\usepackage[dvips]{color}
\usepackage{epsf}
\usepackage{times}
\usepackage{epsfig}
\usepackage{graphicx}
\usepackage{epstopdf}
\usepackage{algorithm}
\usepackage{algorithmic}
\usepackage{amsmath}
\usepackage{amssymb}
\usepackage{amsxtra}
\usepackage{multirow}
\usepackage{mathtools}
\usepackage[font=normalsize]{caption}
\usepackage{cite}	
\usepackage{color}
\usepackage{xcolor}
\usepackage{bbm}
\usepackage[font=small]{caption}
\usepackage{subcaption}
\usepackage{multirow}
\title{Leveraging Personal Navigation Assistant Systems Using Automated Social Media Traffic Reporting}

\author{
\IEEEauthorblockN{\large Xiangpeng Wan, Hakim Ghazzai, and Yehia Massoud}

\IEEEauthorblockA{\small School of Systems and Enterprises, Stevens Institute of Technology, Hoboken, NJ, USA\\ Email: \{xwan6, hghazzai, ymassoud\}@stevens.edu\\} 

{\thanks {\hrule
				\vspace{0.1cm} This paper is accepted for publication in IEEE Technology Engineering Management Society International Conference (TEMSCON'20), Metro Detroit, Michigan (USA).
                
                2020 IEEE. Personal use of this material is permitted. Permission from
                IEEE must be obtained for all other uses, in any current or future media,including reprinting/republishing this material for advertising or promotional purposes, creating new collective works, for resale or redistribution to servers or lists, or reuse of any copyrighted component of this work in other works.
		}}
}

\begin{document}
\maketitle
\begin{abstract}
\boldmath
Modern urbanization is demanding smarter technologies to improve a variety of applications in intelligent transportation systems to relieve the increasing amount of vehicular traffic congestion and incidents. Existing incident detection techniques are limited to the use of sensors in the transportation network and hang on human-inputs. Despite of its data abundance, social media is not well-exploited in such context. In this paper, we develop an automated traffic alert system based on Natural Language Processing (NLP) that filters this flood of information and extract important traffic-related bullets. To this end, we employ the fine-tuning Bidirectional Encoder Representations from Transformers (BERT) language embedding model to filter the related traffic information from social media. Then, we apply a question-answering model to extract necessary information characterizing the report event such as its exact location, occurrence time, and nature of the events. We demonstrate the adopted NLP approaches outperform other existing approach and, after effectively training them, we focus on real-world situation and show how the developed approach can, in real-time, extract  traffic-related information and automatically convert them into alerts for navigation assistance applications such as navigation apps. 
\end{abstract}
\begin{IEEEkeywords}
Intelligent transportation system, natural language processing, social media, Question-Answering model, navigation assistance.
\end{IEEEkeywords}

\section{Introduction}
The increasing amount of vehicular traffic  
is causing severe congestion in many cities all over the world. For instance, according to the 2019 mobility report issued by the New York State Department of Transportation (DOT), the average midtown core speed is consistently 30\% slower than the one of the central business district of Manhattan and, in general, the average speed has decreased by 20\% from 2010 to 2019~\cite{DOT2019}. Traffic congestion becomes a serious problem in modern urban cities due to its social, economic, and environmental impacts. According to a study of the Partnership for New York City, the New York City economy is expected to lose 20 billion a year and more than 100 billion over the next five years due to traffic congestion~\cite{NBJ2018} due to, for example, extra business expenses, excessive fuel consumption, and additional vehicle operating cost. 
Traffic incidents such as unexpected roadside constructions or car accidents, as well as unsynchronized traffic signals represent one of the major causes of traffic congestion in urban areas~\cite{opinionarticle2013}. 
Some novel technologies have been proposed to cope with this problem such as employing smart traffic signals to adapt its time to the traffic flow or installing roadside sensors/cameras to monitor traffic flow and collect data for local authorities for better traffic assessment. Nevertheless, such techniques are exclusively controlled by a single entity and do not involve the participation of road users to combat traffic congestion. In fact, the authors of~\cite{dia2011development} have indicated that every minute of unknown traffic incidents would result in four minutes traffic delay. Therefore, modern technical solutions are proposed to inform drivers the traffic situation in real-time, which can help them avoid congested roads in advance~\cite{8836809,8843858}. For example, Google and Waze proposed human input functionalities where road users can report incidents and share traffic information through the App., and hence, drivers can be informed about the traffic situation. However, these reporting solutions are depending on the direct involvement of road users, who may not participate in the reporting process at all or even if they participate, their inputs might be inaccurate or erroneous. Moreover, it is usually unsafe and uncomfortable for drivers to report incidents while driving. The emergence of Vehicular Social Networks (VSN) represent another potential solution that can be effectively exploited to automate the share of traffic data among vehicles using vehicular communication technologies and social networks and hence, reduce the driver intervention~\cite{7926916,electronics9040648}. 

Social media is a promising source of data that has dramatically evolved over the past decade. People are now getting used to post and massively share information about everything, including transportation. In fact, there are more than 330 million monthly active users on Twitter alone~\cite{Twitter2019}. This generates more than 600 million messages per day. Therefore, social media can be another source of massive traffic data information that can be used to complement existing technologies to get real-time traffic data to be shared among drivers in a given area of interest. Currently, tremendous traffic related information are posted in social media by specialized traffic agencies and regular users, but many of them are not checked by road users, especially drivers who cannot read them while driving. Mining social media data to extract information is a cost-effective way compared to the existing methods where traffic data is mostly collected by physical sensors and authority agencies~\cite{8836726}. More importantly, users tend to post about the traffic conditions they are currently facing by texting when they are caught in traffic jams or by speaking to a voice assistant while driving with Artificial Intelligent (AI) Virtual Assistant~\cite{8813584}. This instant feature of sharing information in social media provides faster alerts and covers ubiquitous locations compared to existing methods~\cite{7359138}. However, the challenges remain in obtaining desired information from millions of social media posts. First, many of the messages posted on social media are usually short, written in an informal language with multiple grammatical errors and misspellings. Secondly, there is only a small percentage of messages related to the real-time traffic status. Lastly, even if we successfully filter the messages and collect traffic related data, we still need to conduct an explicit process to extract the required information, including when and where the incidents happened, what is the expected delay, and which road/lane is blocked, etc. Therefore, it is very worthwhile to provide an efficient text analyzer to collect, filter, and extract essential information from the social media. 
\begin{figure*}[t!]
	\vspace{0.2cm}
	\centerline{\fbox{\includegraphics[height=7cm,width=17cm]{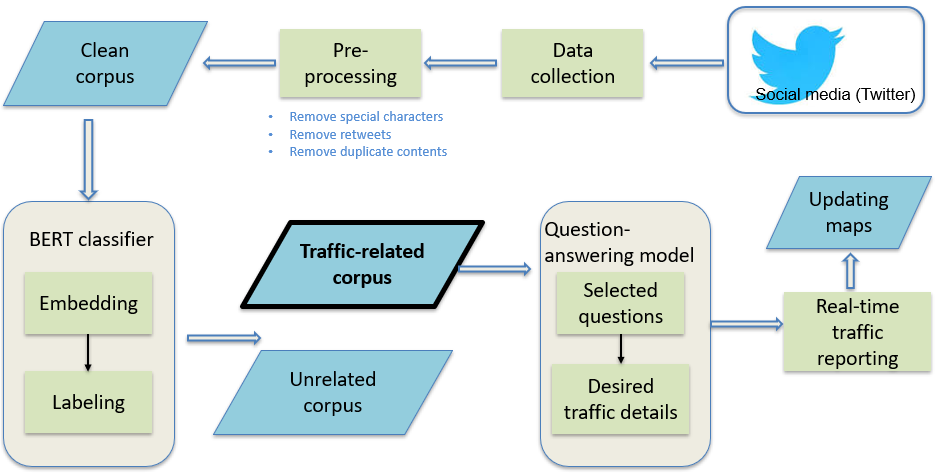}}}\vspace{-0.1cm}
	\caption{\, The proposed navigation assistance framework using social media input.   \normalsize}\label{Fig1}
	\vspace{-0.3cm}
\end{figure*}

In this paper, we propose a joint text processing framework  based on fine-tuning BERT binary classification and the Question-answering models that automatically extracts real-time traffic information data from social media. The proposed framework consists of two phases. In the first phase, we develop a filter that classifies numerous collected text inputs into two groups: traffic-related and traffic-unrelated text. Hence, once successfully trained, the framework can automatically detect traffic-related messages or posts to be fed to the second phase. The latter phase is developed to extract, from these detected traffic-related text, the necessary information to understand and characterize the reported traffic event by determining its location, its occurrence time, and its nature, e.g., blocked road, accident, etc. To do so, we implement the question answering (QA) model~\cite{kumar2016ask}, which is able to automatically generated accurate answers when properly asked questions are used. In fact, recently published models including BERT and XLNet~\cite{zhang2019sg} achieve similar performance compared to human in this challenging area when tested on the Stanford Question Answering Dataset~\cite{rajpurkar2016squad}. Once the classification and question-answering models are developed and trained , we implement an automated system that  converts collected posts from social media into real-time traffic information that are used to update navigation maps and warn drivers about any traffic events. The proposed framework is developed to complement existing navigation approach by empowering navigation assistance system with the power of social networks. Our analysis are based on experiments conducted on Twitter where real-world traffic events are studied. 

The rest of the paper is organized as follows. Section II lists the related work. The objective of the framework and the adopted methodology are presented in Section III. In Section IV, we present the implementation of the proposed framework as well as some performance analysis of the developed models. Finally, conclusions are drawn in Section V.

\section{Related Work}
There are some existing solutions to deal with and assess traffic congestion in ITS. In \cite{hussain2013vehicle}, the authors introduced a framework to collect accident evidence from vehicular sensors to provide a new data source about traffic behavior supervision. In \cite{yan2017short}, the authors predicted short-term traffic conditions with trajectory data using Support Vector Machine which classified the traffic conditions into smooth, basically smooth, mild congestion, moderate congestion, and serious congestion groups. Those categories are used to predict the future traffic situation, however, the accuracy of this method is not well validated. 
\textcolor{black}{Many social media-powered ITS applications exist in the literature. The authors of \cite{8656974} developed an unsupervised learning model for clustering areas based on \textit{placeness} - the cultural and semantic characteristics of a location. In \cite{8720995}, the authors classify the congestion from Twitter and Waze Map Using Artificial Neural Network. The authors of \cite{7732281} explore applying data mining methods to predict traffic congestion and movement patterns from social media posts.
}

With the emergence of NLP techniques, some researchers have explored multiple models to improve classification accuracy and flexibility. In~\cite{booth2015robust}, the authors present a natural language interface for trip planning in complex multi-modal urban transportation networks and provide robust understanding of complex requests while giving users
flexibility in their languages. \textcolor{black}{The model's performance suffers from a lack of flexibility and too little training data.}
On the other hand, word embedding combined with machine learning algorithms has become a widely used technique which computes continuous vector representations of words from very large data sets~\cite{mikolov2013efficient}. The authors of~\cite{8317967} processed a Support Vector Machine algorithm (SVM) based on word vector features that can detect traffic-related tweets with an accuracy of 88.28\%. The authors of~\cite{8527675} proposed \textcolor{black}{Convolutional Neural Network (CNN), Long Short-Term Memory Neural Network (LSTM), and hybrid (LSTM-CNN) models that are trained to classify micro blog posts related to traffic. They found their approaches achieved higher than 91\% accuracy}. In recent years, there is a new trend in designing NLP models after the introduction of transformer model. Recent start-of-the-art NLP models are usually designed as the combination of large pre-trained word-embedding with the transformer model structure. A successful example is the Bidirectional Encoder Representations from Transformers (BERT), developed by Google in 2018, which redefines the start-of-the-art for eleven NLP tasks~\cite{devlin2018bert}. In~\cite{8864964}, the authors explored the BERT to verify that its context-aware representation can achieve similar performance improvement in sentiment analysis. They indicated that the accuracy of classifying traffic-related and traffic-unrelated messages could also be improved by applying BERT.

\textcolor{black}{From our review, we see that the accuracy of classification models related to traffic prediction can be improved. Applying NLP seems to be the most promising direction to gain improvements over previous work. Moreover, the problem of assisting drivers' navigation by exploiting massive social media input has been slightly investigated in the literature. }

\section{Objectives and Methodology}
\label{Sec2}
We propose to design a novel social media data filter and analysis framework that can be used to assist drivers in their navigation and assess the traffic situation beforehand . It consists of two major parts: one part is to filter posts and categorize them into related and unrelated traffic information and the other part is to extract the required information so it can be converted into precise traffic incident events, e.g., to be used by a navigation assistant. The proposed framework is proposed to exploit the power of massive data available in social media and complement existing navigation solutions.
\begin{figure}[t!]
	\centerline{\fbox{\includegraphics[width=8cm]{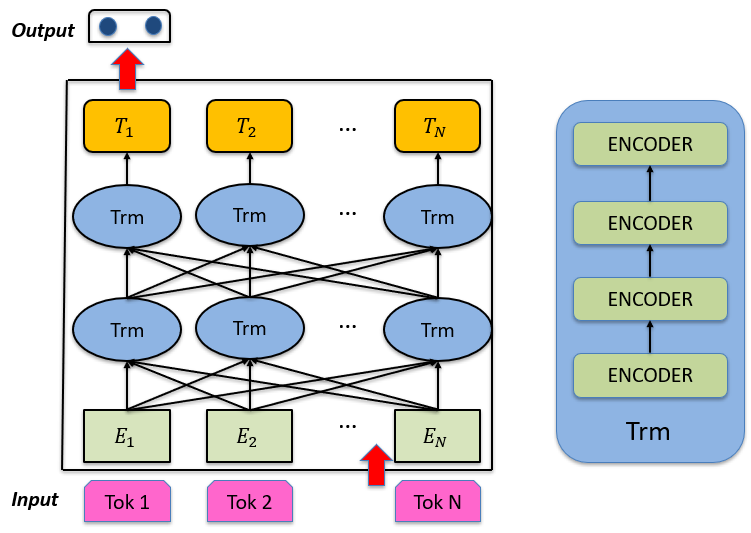}}}\vspace{-0.1cm}
	\caption{\,Architecture of the BERT classification model. \normalsize}\label{Fig2}
	\vspace{-0.5cm}
\end{figure}

The overview of the proposed social media-based framework for navigation assistance is given in Fig.~1. The input of the proposed framework is a raw data of text posts, e.g., tweets. The process starts by a data cleaning procedure that removes special characters and duplicated contents. Then, the fine-tuning BERT classification model is employed to split data into traffic-related and traffic-unrelated corpuses. Afterwards, the traffic-related corpus is fed into the question-answering model to extract required and important information. With dedicatedly selected questions, the model is able to extract desired and necessary traffic information such as where and when the incident exists and what is the expected delay, etc. Finally, the framework updates the navigation maps using these information in an automated manner. 
\begin{figure}[t!]
	\centerline{\fbox{\includegraphics[width=8cm, height=6.5cm]{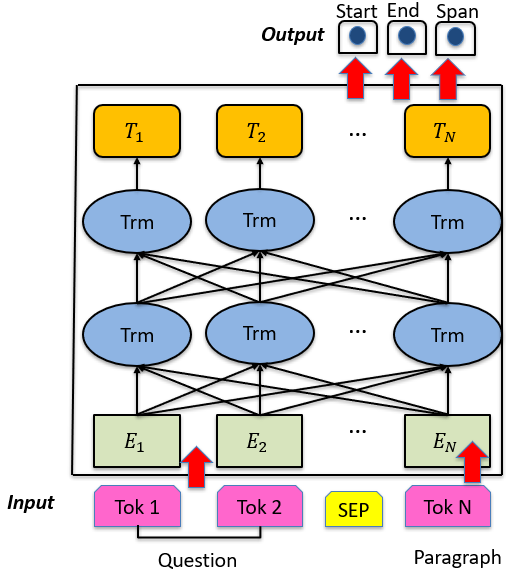}}}\vspace{-0.1cm}
	\caption{\,Architecture of the BERT question-answering model.   \normalsize}\label{Fig3}
	\vspace{-0.5cm}
\end{figure}

In order to filter the tweets to get the traffic-related corpus, we use the fine-tuning BERT classification model, a pre-trained deep bidirectional transformer model for language understanding tasks. It has been shown that the BERT model is achieving extraordinary performance in many tasks such as \textcolor{black}{named entity recognition (NER) and question answering (QA) via transfer learning - BERT is a pre-trained model that can be extended to other NLP problems by adding an output layer at the end of the network\cite{devlin2018bert}}. Indeed, BERT relies on a transformer architecture which contains several self-attention encoders to read the text input. The model architecture is shown in Fig.~\ref{Fig2}. The box $E_n$ where $n=1,\dots,N$ represents the sequence of tokens converted by the input text using WordPiece embedding~\cite{wu2016google}, which would be processed in the neural network later. The output boxes $T_n$ represents the final hidden vector. The first token of every sequence is always a special classification token, $E_1$ in this case. The final hidden state corresponding to this token is used for the classification tasks, $T_1$ in this case. The model is pre-trained on two separate tasks: predicting masked tokens and predicting the next sentence. As a result, the model with the pre-trained parameters could be used for a wide variety of fine-tuning tasks that beat previous proposed NLP models. On the other hand, pre-training and fine-tuning procedures of the model are using the same architecture but different output layer. All parameters are fine-tuned during the fine-tuning process, where there are total 340 million parameters in $BERT_{LARGE}$ models. In fact, fine-tuning is relatively inexpensive compared to the pre-training process. It usually takes few hours on a Graphics Processing Unit (GPU).

With the classification model, we are able to filter the traffic-related information, then we apply question-answering model to extract desired traffic information from the traffic-related corpus. The question-answering model that we used, shown in Fig~\ref{Fig3}, is also the fine-tuning BERT, with similar architecture as the classification model but with one difference: the output layer. The input contains two parts that are separated by the special character ($SEP$), one is the question and the other is the paragraph where to find the corresponding answers, in our case, a tweet. The output layer contains three neurons: ``start'', ``end'', and ''Span''. The first two neurons indicate the locations of the answer tokens in the tweet and the last neuron represents the total length of the answer, e.g., number of words. With properly selected questions, we are able to automatically extract the desired traffic information from the collected tweets and then, use them for navigation in real-time by, for example, automatically updating the navigation maps.
\begin{table}[b!]
\caption{Examples of labeled tweets in the data set.}
\centering
\begin{tabular}{ |p{6cm}||p{1.2cm}|  }
 \hline
 Tweet & Label \\
 \hline
 \hline
 Crash investigation and Crash with Injuries on Garden State Parkway southbound North of Exit 63B - NJ 72 West (Stafford Twp)  right lane blocked.   & Related \\
 \hline
 Incident on I295 NB at Exit 61 - Arena Dr. &   Related  \\
 \hline
 Cleared: Construction on West 11Th Street from 7th Avenue to 6th Avenue &   Related  \\
 \hline
 I will visit the transportation office tomorrow at Garden Parkway but my car is crashed. &   Unrelated  \\
 \hline
 Can confirm former 
 @lakers star Rick Fox WAS NOT among the passengers on the helicopter with Kobe Bryant. Source: his daughter. & Unrelated \\
 \hline
  Can confirm former 
 We had car troubles along the jos-Abuja road this morning, the army officers at the military check point close to where the incident happened went beyond the call of duty to help, one literally went under the vehicle to try and fix it. & Unrelated \\
 \hline
\end{tabular}
\label{t1}
\end{table} 

\section{Implementation}
\label{Sec3}
In this section, we illustrate and evaluate our proposed system by analyzing recently published tweets data from Twitter that are written in English, and we conduct an experiment by updating real-time traffic navigation map. The entire pipeline is implemented as previously stated in Section III. 

\subsection{Data Collection}
\label{Sec3A}
We start by feeding a large amount of labeled tweets into the classification model to train the parameters, i.e., the weights of the BERT classification model. \textcolor{black}{In order to get a high accuracy, our model must be able to identify the context of the tweets. In other words, it must be able to learn if tweets are related to traffic incidents if no traffic-related terms are used, and it must be able to learn if tweets are unrelated to traffic if they do contain traffic-related terms.} Some examples are given in Table~\ref{t1}. 

\begin{figure}[t!]
	\vspace{0.2cm}
	\centerline{\fbox{\includegraphics[width=8cm]{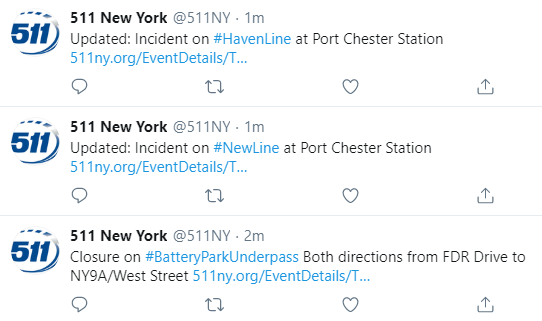}}}\vspace{-0.1cm}
	\caption{\,One example of New York city traffic account on Twitter.   \normalsize}\label{Fig4}
	\vspace{-0.3cm}
\end{figure}

We collect 106437 tweets using Twitter API. Some of them from traffic management agencies and other form regular users of Twitters. Afterward, we categorize them into three major groups: 1) Related tweets that are labeled as one, 2) Unrelated tweets that contain high-frequency traffic words labeled as zero, and 3) Unrelated tweets with no high-frequency traffic words labeled as zero. For training purposes, the number of tweets labeled as one is almost the same as the number of tweets labeled as zero. In order to guarantee that the related tweets are accurately labelled, we crawled the contents from several government official twitter accounts and check them manually. One of the official twitter accounts example is shown in Fig~\ref{Fig4}. Those kinds of twitter accounts exist for some cities which are responsible for updating people about traffic and transit situations.
\begin{figure}[t!]
	\vspace{0.1cm}
	\centerline{\includegraphics[width=9cm]{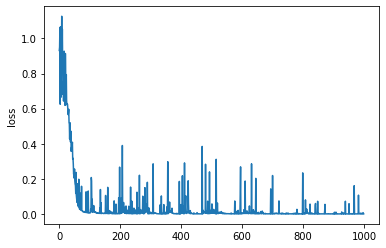}}\vspace{-0.1cm}
	\caption{\, Training phase of the classification model: Achieved loss vs. training steps.   \normalsize}\label{Fig5}
	\vspace{-0.3cm}
\end{figure}
\begin{figure*}[t!]
	\vspace{0.1cm}
	\centerline{\includegraphics[width=17cm, height=6cm]{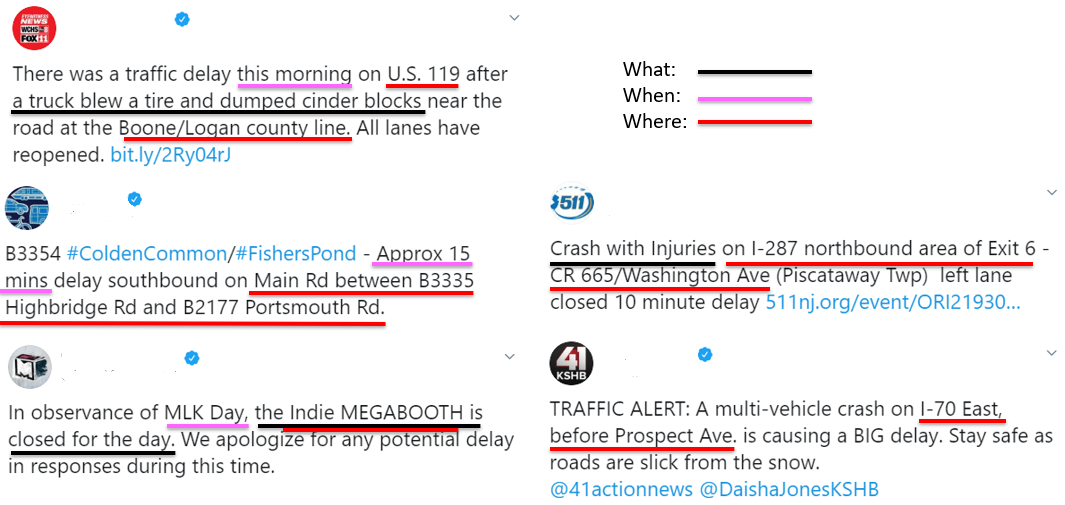}}\vspace{-0.1cm}
	\caption{\,Examples of the desired traffic information extracted using the question-answering model. \normalsize}\label{Fig6}
	\vspace{-0.4cm}
\end{figure*}

\textcolor{black}{In order to generalize the predictive power of our model, we inserted random sentences unrelated to traffic in our traffic-related tweets.}

To obtain the second group of tweets, we browse the dataset and search for high-frequency traffic words such as ``incident, delay, construction, crash, lanes, road, etc.'' in each tweet and assign to it the label zero if only few words are overlapping (three and less). Otherwise, the tweet is assigned to the first group. For the last group, we simply select the tweets that do not contain any of the high-frequency traffic words. In our study, the high-frequency traffic word contains 50 words.

\subsection{Evaluation Metrics}
The standard way to evaluate the performance of a given classifier is using the confusion matrix, which contains True Positive (TP), False Negative (FN), True Negative (TN), and False Positive (FP).
The objective is to test the efficiency of the BERT classifier applied our dataset to ensure effective classification of new tweets so later, we can extract desired traffic information from these related tweets.
We define the precision as $\frac{TP}{TP+FP}$, the recall as $\frac{TP}{TP+FN}$, the accuracy as $\frac{TP+TN}{TP+FP+TN+FN}$, and the Matthews correlation coefficient (MCC) as $\frac{TP \times TN - FP \times FN }{\sqrt{(TP+FP)(TP+FN)(TN+FP)(TN+FN)}}$.

\subsection{Classification Results}
To measure the performance of the proposed classification model, we split the total 106437 tweets into two groups: training and testing datasets, which contains 80\% and 20\% of the total tweets, respectively. We set the learning rate to $2 \times 10^{-5}$, batch size to $24$, the optimization steps to $4018$, and the dropout probability to $0.1$. The activation function is Gelu and we train the model with one epoch. In Fig.~\ref{Fig5}, we illustrate the training results where we plot the loss with respect to the training steps. \textcolor{black}{We see from Fig.~\ref{Fig5} that our model training loss function descended rapidly in the first 200 steps, then stabilized afterward.}

\textcolor{black}{Our training results are summarized in Table~\ref{t2}. We compare the performance of our model with Naive Bayes and Support Vector Machine (SVM) - our model has the best Precision, recall, accuracy, and MCC scores.} 
\begin{table}[t!]
\caption{Testing phase: classification results}
\centering
\begin{tabular}{ |c||c||c||c||c|  }
 \hline
 Class & Precision & Recall & Accuracy & MCC\\
 \hline
 \hline
 Naive Bayes & 95.6  &  92.7 & 94.45 & 88.9\\
 \hline
 SVM & 99.0 & 94.3 & 96.8 &93.7 \\
 \hline
 Fine-tuning BERT & 99.6 & 99.3 & 99.45 & 98.9  \\
 \hline
\end{tabular}
\label{t2}
\vspace{-0.3cm}
\end{table} 

\subsection{Question-answering model}
Since we have developed a trustful classification model, we then use it to filter the 10,000 recently published tweet. The classifier detected 148 traffic related tweets with TP $= 137$ and FP $= 11$. Our purpose is to extract the desired traffic information from these tweets by using appropriate question-answering model. To extract the details, we apply the ``Bert -For-Question-Answering'', which is trained on the SQUAD dataset~\cite{wolf2019transformers} using three types of questions: ``What'', ``When'', and ``Where'' to identify the type of incident, its occurrence time, and its location. After training the model for $2$ epochs with a batch size of $8$, the model achieves a matching score of exactly 82.1\% and a F1-score 85.0\% in the testing dataset where human performance is shown to be 86.8\% and 89.5\% for the matching and F1 scores, respectively. With a fast training time, the adopted question-answering model is shown to achieve close performance to the one achieved by humans. 
Examples showing how the question-answering model extracts desired traffic details are given in Fig~\ref{Fig6}.

We notice that, with the intentionally selected questions mentioned earlier, we are able to extract the exact locations of the happening incidents. We also could extract a brief information about what happened exactly in that location, i.e., the type of the incident. From all 137 traffic related tweets, we can extract the desired traffic information from 118 of them, where the percentage is 86\%, meets our expectation. Although the desired traffic details could be extracted, it is not straightforward to automatically handle the information. For instance, the exact location at which the incident occurred could be a short phrase with several road names, e.g., ``Main Rd between B3335 Highbridge Rd and B2177 Portsmouth Rd''. Our solution is to create a comprehensive word document that includes terms such as``and, between, and of, etc.'' so that we could split the short sentences into different phrases and each phrase represents a specific location. In other words, we could split ``Main Rd between B3335 Highbridge Rd and B2177 Portsmouth Rd'' to ``Main Rd'',``B3335 Highbridge Rd'', and ``B2177 Portsmouth Rd''. 

\begin{figure}[t!]
	\vspace{0.2cm}
	\centerline{\includegraphics[width=9cm,height=8cm]{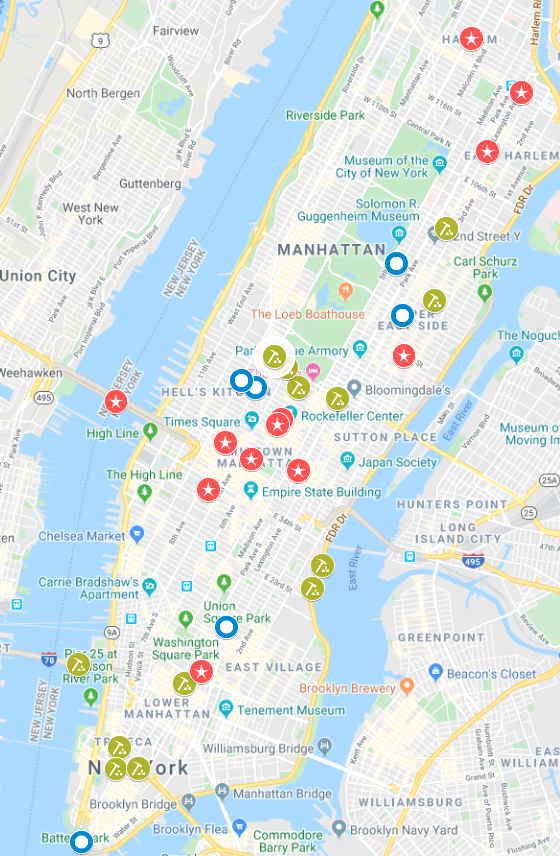}}\vspace{-0.1cm}
	\caption{\,Example of an updated map over a five hours period. The map is only updated using information extracted from the collected tweets. The red star represents an accident, the green axe represents a construction, and the blue circle represents a temporally closure.   \normalsize}\label{Fig7}
	\vspace{-0.4cm}
\end{figure}

On the other hand, the relations between those specific locations are essential to be considered. For example, if the relation is ``between'', then we must update the delayed routes between the two specific locations and potentially other affected routes. In the previous example, the ``Main Rd'' between ``B3335 Highbridge Rd'' and ``B2177 Portsmouth Rd'', is the affected route. If the relation is ``of'', like in ``I-287 northbound area of Exit 6 - CR 665/Washington Ave'' then, we need to identify the delayed routes as ``I-287 northbound area'' in the region of ``Exit 6 - CR 665/Washington Ave''. In other words, our designed automated system needs to handle these issues and convert them into realistic and accurate data in the navigation map. To validate our system, we use the recently published tweets from Manhattan area, New York City, to draw a real-time traffic map, as shown in Fig~\ref{Fig7}.

\section{Conclusion}
\label{Sec4}
In this paper, we proposed an automated traffic information extractor framework. From a massive input of social media data, the developed NLP-based framework conducts a filtering process to distinguish traffic-related posts from others. Then, a question-answering model is applied to characterize the reported events and extract key information about them. This allows the rapid and automated integration of such alerts into navigation platforms, which help drivers assess the traffic situation beforehand and avoid congested roads. The performance of the proposed framework has been experimented on the area of Manhattan, New York City, where real-time traffic maps are automatically updated using information extracted by scrapping Twitter.

\bibliographystyle{ieeetr}
\bibliography{main}

\begin{thebibliography}{10}

\bibitem{DOT2019}
{NYC Department of Transportation}, ``New york city mobility report 2019,''
  tech. rep., Aug. 2019.

\bibitem{NBJ2018}
{A. Noto}, ``{NYC} economy may be losing 20 billion a year due to traffic
  congestion,'' tech. rep., Jan. 2018.

\bibitem{opinionarticle2013}
{A. Rosen}, ``What really causes traffic congestion?,'' tech. rep., JUL. 2013.

\bibitem{dia2011development}
H.~Dia and K.~Thomas, ``Development and evaluation of arterial incident
  detection models using fusion of simulated probe vehicle and loop detector
  data,'' {\em Information Fusion}, vol.~12, no.~1, pp.~20--27, 2011.

\bibitem{8836809}
X.~{Wan}, H.~{Ghazzai}, and Y.~{Massoud}, ``Real-time navigation in urban areas
  using mobile crowd-sourced data,'' in {\em IEEE International Systems
  Conference (SysCon'19)}, Orlando, FL, USA, Apr. 2019.

\bibitem{8843858}
X.~{Wan}, H.~{Ghazzai}, and Y.~{Massoud}, ``Mobile crowdsourcing for
  intelligent transportation systems: Real-time navigation in urban areas,''
  {\em IEEE Access}, vol.~7, pp.~136995--137009, 2019.

\bibitem{7926916}
Z.~{Ning}, F.~{Xia}, N.~{Ullah}, X.~{Kong}, and X.~{Hu}, ``Vehicular social
  networks: Enabling smart mobility,'' {\em IEEE Communications Magazine},
  vol.~55, pp.~16--55, May 2017.

\bibitem{electronics9040648}
X.~Wan, H.~Ghazzai, and Y.~Massoud, ``A generic data-driven recommendation
  system for large-scale regular and ride-hailing taxi services,'' {\em
  Electronics}, vol.~9, no.~4, 2020.

\bibitem{Twitter2019}
{ J. Clement}, ``Twitter: number of monthly active {U.S.} users 2010-2019,''
  tech. rep., Aug. 2019.

\bibitem{8836726}
M.~C. {Lucic}, H.~{Ghazzai}, and Y.~{Massoud}, ``A generalized and dynamic
  framework for solar-powered roadside transmitter unit planning,'' in {\em
  IEEE International Systems Conference (SysCon'19)}, Orlando, FL, USA, Apr.
  2019.

\bibitem{8813584}
A.~P. {Sam}, B.~{Singh}, and A.~S. {Das}, ``A robust methodology for building
  an artificial intelligent (ai) virtual assistant for payment processing,'' in
  {\em 2019 IEEE Technology Engineering Management Conference (TEMSCON)},
  pp.~1--6, 2019.

\bibitem{7359138}
X.~{Zheng}, W.~{Chen}, P.~{Wang}, D.~{Shen}, S.~{Chen}, X.~{Wang}, Q.~{Zhang},
  and L.~{Yang}, ``Big data for social transportation,'' {\em IEEE Transactions
  on Intelligent Transportation Systems}, vol.~17, pp.~620--630, March 2016.

\bibitem{kumar2016ask}
A.~Kumar, O.~Irsoy, P.~Ondruska, M.~Iyyer, J.~Bradbury, I.~Gulrajani, V.~Zhong,
  R.~Paulus, and R.~Socher, ``Ask me anything: Dynamic memory networks for
  natural language processing,'' in {\em International conference on machine
  learning}, pp.~1378--1387, 2016.

\bibitem{zhang2019sg}
Z.~Zhang, Y.~Wu, J.~Zhou, S.~Duan, and H.~Zhao, ``{SG-Net}: Syntax-guided
  machine reading comprehension,'' {\em arXiv preprint arXiv:1908.05147}, 2019.

\bibitem{rajpurkar2016squad}
P.~Rajpurkar, J.~Zhang, K.~Lopyrev, and P.~Liang, ``Squad: 100,000+ questions
  for machine comprehension of text,'' {\em arXiv preprint arXiv:1606.05250},
  2016.

\bibitem{hussain2013vehicle}
R.~Hussain, F.~Abbas, J.~Son, D.~Kim, S.~Kim, and H.~Oh, ``Vehicle witnesses as
  a service: Leveraging vehicles as witnesses on the road in vanet clouds,'' in
  {\em IEEE International Conference on Cloud Computing Technology and Science
  (CloudCom 2013)}, Bristol, UK, Mar. 2013.

\bibitem{yan2017short}
H.~Yan and D.-J. Yu, ``Short-term traffic condition prediction of urban road
  network based on improved {SVM},'' in {\em IEEE International Smart Cities
  Conference (ISC2 2017)}, Wuxi, China, Sept. 2017.

\bibitem{8656974}
G.~K. et~al., ``Location digest: A placeness service to discover community
  experience using social media,'' in {\em IEEE International Smart Cities
  Conference (ISC2'18)}, Kansas City, MO, USA, Sept. 2018.

\bibitem{8720995}
``Congestion correlation and classification from twitter and waze map using
  artificial neural network,'' in {\em 2018 3rd International Conference on
  Information Technology, Information System and Electrical Engineering
  (ICITISEE)}, pp.~224--229, Nov 2018.

\bibitem{7732281}
H.~{Shekhar}, S.~{Setty}, and U.~{Mudenagudi}, ``Vehicular traffic analysis
  from social media data,'' in {\em 2016 International Conference on Advances
  in Computing, Communications and Informatics (ICACCI)}, pp.~1628--1634, Sep.
  2016.

\bibitem{booth2015robust}
J.~Booth, B.~Di~Eugenio, I.~F. Cruz, and O.~Wolfson, ``Robust natural language
  processing for urban trip planning,'' {\em Applied Artificial Intelligence},
  vol.~29, no.~9, pp.~859--903, 2015.

\bibitem{mikolov2013efficient}
T.~Mikolov, K.~Chen, G.~Corrado, and J.~Dean, ``Efficient estimation of word
  representations in vector space,'' {\em arXiv preprint arXiv:1301.3781},
  2013.

\bibitem{8317967}
A.~{Salas}, P.~{Georgakis}, and Y.~{Petalas}, ``Incident detection using data
  from social media,'' in {\em 2017 IEEE 20th International Conference on
  Intelligent Transportation Systems (ITSC)}, pp.~751--755, Oct 2017.

\bibitem{8527675}
Y.~{Chen}, Y.~{Lv}, X.~{Wang}, L.~{Li}, and F.~{Wang}, ``Detecting traffic
  information from social media texts with deep learning approaches,'' {\em
  IEEE Transactions on Intelligent Transportation Systems}, vol.~20, no.~8,
  pp.~3049--3058, 2019.

\bibitem{devlin2018bert}
J.~Devlin, M.-W. Chang, K.~Lee, and K.~Toutanova, ``{ERT}: Pre-training of deep
  bidirectional transformers for language understanding,'' {\em arXiv preprint
  arXiv:1810.04805}, 2018.

\bibitem{8864964}
Z.~{Gao}, A.~{Feng}, X.~{Song}, and X.~{Wu}, ``Target-dependent sentiment
  classification with {BERT},'' {\em IEEE Access}, vol.~7, pp.~154290--154299,
  2019.

\bibitem{wu2016google}
{Y. Wu, et al.}, ``Google's neural machine translation system: Bridging the gap
  between human and machine translation,'' {\em arXiv preprint
  arXiv:1609.08144}, 2016.

\bibitem{wolf2019transformers}
{T. Wolf et al.}, ``Transformers: State-of-the-art natural language
  processing,'' 2019.

\end{thebibliography}
\end{document}